\documentclass{article}

\newif\ifdashvmcworkshop
\dashvmcworkshoptrue

\PassOptionsToPackage{numbers,compress}{natbib}
\usepackage[dblblindworkshop, final]{neurips_2026}
\workshoptitle{PTA: From Pretrained Representations to Acting Agents}
\newcommand{\dashvmcbibliographystyle}{unsrtnat}

\ifdashvmcworkshop
  \usepackage[utf8]{inputenc}
  \usepackage[T1]{fontenc}
  \usepackage{courier}
  \usepackage{fontawesome5}
\fi
\usepackage{float}
\usepackage{hyperref}
\usepackage{url}
\usepackage{booktabs}
\usepackage{array}
\usepackage{amsfonts}
\usepackage{amsmath}
\usepackage{microtype}
\usepackage{xcolor}
\usepackage{graphicx}
\usepackage{placeins}
\usepackage{flafter}
\usepackage{needspace}

\ifdashvmcworkshop
  \definecolor{citeblue}{HTML}{4A90E2}
  \hypersetup{
    colorlinks=true,
    citecolor=citeblue,
    linkcolor=citeblue,
    urlcolor=citeblue,
    pdftitle={DashVMC: Real-Time Discrete World Model Control in Geometry Dash},
    pdfauthor={Florent Tariolle and Florian Yger}
  }
\else
  \hypersetup{
    pdftitle={DashVMC: Real-Time Discrete World Model Control in Geometry Dash},
    pdfauthor={Florent Tariolle and Florian Yger}
  }
\fi

\newcommand{\ntokens}{64}
\newcommand{\vocabsize}{1000}
\newcommand{\fsqlevels}{[8,5,5,5]}
\newcommand{\gridsize}{8 \times 8}
\newcommand{\imgsize}{64 \times 64}
\newcommand{\ctrlparams}{45{,}546}

\newcommand{\dashvmcabstract}{%
  World-model agents are usually evaluated in simulators that can wait for the policy; live games impose the opposite constraint, requiring capture, prediction, and action before the next frame.
  We present \emph{DashVMC}, which learns a compact, action-conditioned world model from approximately two hours of recorded Geometry Dash gameplay.
  To test whether the learned dynamics are actionable, a controller is initialized by behavioural cloning (BC) and refined with Proximal Policy Optimization (PPO) entirely in frozen-model rollouts, without further interaction with the live game.
  Across three controller seeds, the refined policies survive longer than their BC initializations on all three official levels and a held-out community layout.
  At deployment, the baseline skips visual generation and sustains a 60-Hz capture-to-action loop on a consumer GPU.
  Action-conditioned continuations and rollout diagnostics show that the model remains useful for control despite imperfect long-horizon fidelity.%
}

\newcommand{\dashvmcacknowledgments}{%
  We thank Cl\'ement Chatelain and Robin Condat for their guidance during the Representation Learning course at INSA Rouen Normandy, in which this project began, and CRIANN for providing the computing resources used for training.
  We thank Ma{\"e}l Planchot for contributing gameplay data.
  Geometry Dash is developed by RobTop Games.%
}

\ifdashvmcworkshop
  \title{DashVMC: Real-Time Discrete World Model Control\\in Geometry Dash}
  \author{%
    Florent Tariolle\thanks{Corresponding author: \href{mailto:florent.tariolle@insa-rouen.fr}{\texttt{florent.tariolle@insa-rouen.fr}}.} \\
    INSA Rouen Normandy
    \And
    Florian Yger \\
    INSA Rouen Normandy, LITIS
  }
\else
  \title{%
    DashVMC\\[0.1em]
    {\large\color{black!70}Real-Time Discrete World Model Control in Geometry Dash}%
  }
  
  \abstract{\dashvmcabstract}
  \metadata[Project page]{\href{https://tariolle.github.io/dash-vmc/}{tariolle.github.io/dash-vmc}}
  \metadata[Code]{\href{https://github.com/Tariolle/dash-vmc}{github.com/Tariolle/dash-vmc}}
  \correspondence{\email{florent.tariolle@insa-rouen.fr}}
\fi

\begin{document}

\maketitle

\ifdashvmcworkshop
  \vspace{-0.12in}
  {\centering\hypersetup{urlcolor=black}
    \href{https://tariolle.github.io/dash-vmc/}{\faGlobe~Website}\qquad
    \href{https://github.com/Tariolle/dash-vmc}{\faGithub~Code}\par}
  \vspace{0.1in}

  \begin{abstract}
    \dashvmcabstract
  \end{abstract}
\fi

\section{Introduction}
\label{sec:intro}

A world model learns transition dynamics \(p_\theta(s_{t+1}\mid s_t,a_t)\) and can be rolled forward so that an agent practices without further environment interaction~\citep{lecun2022path,ha2018world}.
DashVMC asks whether such practice can improve a policy that must then act in a live, non-paused game.
It predicts a 64-token next grid in parallel for inexpensive imagined rollouts.
The baseline deployment skips generation and reuses the transformer's temporal summary; a matched ablation tests whether the transformer must remain in the live loop.

Most world-model agents are evaluated in synchronous environments that wait for the policy.
A live, non-paused application reverses that contract: screen capture, perception, inference, and action dispatch must finish before the action becomes stale.

Geometry Dash is a small but unforgiving test of this contract.
The avatar moves continuously, the action space is binary, and a mistimed jump causes immediate death.
Levels are deterministic, but the agent observes only captured pixels and acts through keyboard events; it does not receive emulator state during control.
The apparent simplicity therefore isolates a demanding question: can an agent learn when to act from pixels, improve without touching the live game, and still react within narrow timing windows on modest hardware?

DashVMC follows the perception--dynamics--control organization of World Models~\citep{ha2018world}: it compresses each frame into a discrete grid, learns action-conditioned grid dynamics, initializes a lightweight action model with BC, and refines it with PPO~\citep{schulman2017ppo} in the frozen model.

\needspace{5\baselineskip}
Our contributions are:
\begin{enumerate}
  \item a compact discrete, action-conditioned world model learned from approximately two hours of offline gameplay, whose rollouts support policy refinement without further live-game interaction;
  \item a decoder-free baseline that sustains a 60-Hz capture-to-action loop on a consumer GPU, together with a spatial-only variant that removes the transformer at inference without a consistent live-performance deficit; and
  \item evidence that the learned dynamics are actionable: controllers refined only in frozen-model rollouts improve over their exact BC parents on all three official levels across three seeds and on a held-out layout.
\end{enumerate}
Diagnostics cover action sensitivity, rollout fidelity, controller input, dynamics losses, and latency.

\FloatBarrier
\section{Related work}
\label{sec:related}

\paragraph{World models and imagination.}
World Models demonstrated that a compact controller optimized in learned dreams can transfer to the real environment~\citep{ha2018world}.
IRIS combines a discrete autoencoder with an autoregressive transformer for data-efficient Atari learning~\citep{micheli2023iris}; \(\Delta\)-IRIS reduces sequence cost with discrete delta tokens and continuous context tokens~\citep{micheli2024deltairis}; DIAMOND uses diffusion for imagination-trained control~\citep{alonso2024diamond}; and DreamerV3 scales categorical recurrent latents across domains~\citep{hafner2025dreamerv3}.
TWISTER adds action-conditioned CPC to transformer dynamics~\citep{burchi2025twister}, which DashVMC adopts alongside parallel spatial prediction.
Earlier screenshot-based Geometry Dash control used DQN and imitation under a fabricated timestep~\citep{li2017geometrydash}; here the game clock continues during capture and inference.

\paragraph{FSQ tokenization and policy initialization.}
FSQ replaces learned vector-quantization codebooks with bounded scalar levels, avoiding commitment losses and codebook collapse while exposing an integer coordinate lattice~\citep{mentzer2024fsq}; we use the improved iFSQ bounding function~\citep{lin2026ifsq}.
Demonstration-based RL can reduce costly early exploration~\citep{hester2018deep,baker2022vpt}; here recorded actions initialize the controller, while the retained policy is selected only after imagined PPO refinement.
Deterministic Sobel preprocessing removes appearance detail before tokenization, providing a compact task-specific alternative to richer continuous or generative representations.

\FloatBarrier
\section{DashVMC}
\label{sec:method}

DashVMC is organized around a simple loop: recorded play trains a world model, the action model is refined in rollouts from the frozen dynamics model, and the resulting controller is returned to the live game.
The visual encoder, dynamics model, and action model are trained in sequence and then frozen for deployment.
Figure~\ref{fig:pipeline} shows how the same learned state supports both imagined practice and live reaction.

\ifdashvmcworkshop
  \begin{figure}[!htbp]
\else
  \begin{figure}[!htbp]
\fi
  \centering
  \ifdashvmcworkshop
    \includegraphics[width=\textwidth]{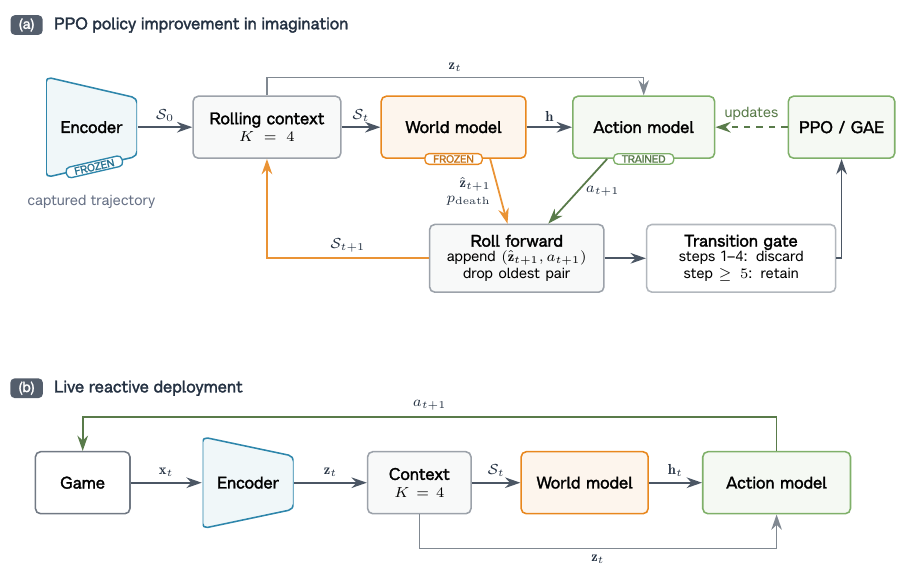}
  \else
    \includegraphics[width=0.82\textwidth]{figures/pipeline.pdf}
  \fi
  \caption{Learning in imagination and acting in the live game. \textbf{(a)} A recorded context seeds the frozen world model; the action model chooses what to do, the world model predicts the consequence, and PPO updates only the action model. \textbf{(b)} In the baseline live path, the encoder and temporal context provide the state used for action selection, while visual generation is skipped.}
  \label{fig:pipeline}
\end{figure}

\subsection{Visual tokenization}
\label{sec:fsq}

Each RGB capture is cropped, converted to grayscale, filtered with Sobel edges, and resized to \(\imgsize\).
A convolutional autoencoder maps the result through three stride-2 stages to a four-channel \(\gridsize\) latent map.
Each spatial vector is independently quantized with iFSQ levels \(\fsqlevels\), using the bounding function \(2\operatorname{sigmoid}(1.6z)-1\), and yields \(\ntokens\) token IDs from \(\vocabsize\) implicit codes~\citep{mentzer2024fsq,lin2026ifsq}.
The tokenizer is trained on adjacent frames with reconstruction, temporal-slowness, and latent-uniformity objectives~\citep{xia2025grwm}.
Once selected, its encoder is frozen; its decoder is used to visualize dreams but is not required by the action model.

\ifdashvmcworkshop\else
  \FloatBarrier
\fi
\subsection{Action-conditioned dynamics}
\label{sec:transformer}

The dynamics model is an eight-layer, width-384 transformer that receives four recent frame--action pairs.
Each frame block contains 64 visual tokens and one \textsc{alive}/\textsc{death} status token, with action tokens inserted between blocks; a final masked block requests the next frame.
Attention is bidirectional within a frame and causal across frame/action blocks, so all next-frame positions are predicted in one forward pass without seeing their targets.
An explicit action token separates consecutive frames, so the model can answer the question central to policy learning: what changes if the action model jumps rather than idles?
It predicts all 64 codes of the next grid in parallel, together with an \textsc{alive}/\textsc{death} score, and exposes a temporal summary \(\mathbf h_t\), taken from the most recent action-token position, to the action model.
Parallel prediction keeps imagined steps inexpensive.

Training combines next-grid prediction with action-conditional contrastive predictive coding of weight 0.1~\citep{burchi2025twister,oord2018cpc}, context corruption, and an auxiliary death objective.
For context corruption, 5\% of context tokens are replaced uniformly and another 5\% with adjacent FSQ codes.
The token loss uses focal modulation~\citep{lin2017focal} and structured label smoothing (SLS): for target lattice coordinate \(\mathbf c_i\), smoothing mass \(\varepsilon=0.1\) is distributed over non-target codes proportionally to \(\exp(-\|\mathbf c_i-\mathbf c_j\|_2^2/2\sigma^2)\), with \(\sigma=0.9\).
Focal modulation is applied once to the aggregate soft-target cross-entropy, and the output projection is tied to the token embedding.

\subsection{Controller and latent policy improvement}
\label{sec:controller}

The \(\ctrlparams\)-parameter actor--critic embeds the current token grid, processes it with two convolutions and max pooling, and concatenates the resulting 256-dimensional feature with the dynamics model's temporal summary \(\mathbf h_t\).
Separate heads predict jump probability, value, and an eight-action auxiliary output covering the current and next seven actions.
The matched spatial-only ablation removes \(\mathbf h_t\) from BC, PPO, and deployment, leaving a 41,706-parameter action model; the world model still generates its PPO rollouts.
BC first reproduces recorded action timing; PPO then departs from imitation through repeated imagined consequences.

Each PPO iteration generates 512 imagined episodes of at most 45 steps from recorded four-frame contexts, with greedy next-grid predictions and sampled Bernoulli actions.
The first four generated transitions refresh the context but are excluded because the recorded approach may make collision unavoidable; the remainder covers visible obstacles before invented geometry dominates.
Rollouts terminate when the death logit exceeds the \textsc{alive} logit; because death frames are oversampled \(5\times\), this is an operational boundary rather than a calibrated probability.
Reward is \(+1\) per survived step and \(-0.25\) per jump, discouraging a local optimum in which an early jump merely delays collision while airborne and earns extra survival reward without clearing the obstacle.
We use Generalized Advantage Estimation (GAE)~\citep{schulman2015gae} with \(\gamma=0.995\), \(\lambda=0.95\), PPO clip 0.2, and auxiliary-action loss weight 0.1.

\subsection{Reactive deployment path}
\label{sec:deployment-method}

Training generates consequences; baseline live control only interprets recent observations.
Desktop Duplication captures only the \(1032\times1032\) model crop; CPU grayscale conversion followed by CUDA Sobel filtering and area resizing produces the \(64\times64\) observation.
Each frame is encoded and appended to a GPU-resident context, and the transformer runs context prefill only to produce \(\mathbf h_t\).
The live context marks every observation \textsc{alive}; process-memory death detection only terminates and restarts an attempt outside the controller.
The next-grid predictor and image decoder are skipped, reducing the deployed path to 15.57 million parameters; the encoder uses \texttt{torch.compile}, and transformer context encoding uses a CUDA Graph after warm-up.
The accelerated preprocessing path was checked against the recorded-data pipeline on synthetic and fresh live captures and produced byte-identical observations.
Spatial-only deployment loads only the frozen encoder and action model; both paths are reactive controllers whose policy improvement occurred during offline imagined training.

\FloatBarrier
\section{Experimental setup}
\label{sec:setup}

\paragraph{Data and training.}
Before augmentation, the corpus contains 4,264 action-labelled gameplay episodes, or 251,963 frames and approximately 2.33 hours at the nominal 30-FPS capture cadence.
It includes deliberate failures, official Levels 1--7, several community levels, and 36 expert trajectories or segments totaling less than 20 minutes.
BC uses action labels from both the deliberate-failure and expert corpora.
Recording samples the current key state after each capture; deployment appends the new capture with the previously held state before dispatching the next, while process memory only segments deaths and restarts.
Mechanics outside the selected scope, including gravity inversion, are excluded.
Failure and expert trajectories are each split 90/10 with seed 42 at the base-episode level.
During tokenizer training, each adjacent-frame pair receives a shared two-dimensional translation whose horizontal and vertical offsets are sampled independently and uniformly from \([-4,4]\) pixels.
For dynamics training, the frozen encoder instead tokenizes each base episode at vertical offsets \(\{-4,-2,0,2,4\}\), expanding the corpus to 21,320 episode variants.
All windows and variants inherit their base-episode assignment; although no complete trajectories are duplicated, deterministic repeated attempts can share near-identical prefixes, so the 10\% stratum is an in-distribution development set rather than a generalization test.
The tokenizer, dynamics model, and BC controller use this split; PPO draws rollout seeds from the full corpus and uses 512 fixed development contexts for checkpoint selection.
Tokenizer and dynamics training use seed 42, while controller training is repeated with seeds 43--45 under the same frozen representation and world model.
The first two stages run sequentially in BF16 on one A100, controller training runs in BF16 on one H200, and live inference and local latency evaluation use the same RTX 2060 with PyTorch 2.11.0+cu126 (CUDA 12.6).

\begin{table}[!htbp]
  \centering
  \footnotesize
  \setlength{\tabcolsep}{4pt}
  \caption{Training and checkpoint selection; tokenizer/dynamics learning rates use cosine decay.}
  \label{tab:training}
  \begin{tabular}{>{\raggedright\arraybackslash}p{0.12\textwidth}>{\raggedright\arraybackslash}p{0.58\textwidth}>{\raggedright\arraybackslash}p{0.18\textwidth}}
    \toprule
    Stage & Optimization & Selection \\
    \midrule
    Tokenizer & Adam; 1,000 epochs; batch 2,048; LR \(10^{-3}\!\to\!10^{-5}\); slowness 0.1; uniformity 0.01. & Min. loss (epoch 920). \\
    Dynamics & AdamW; 200\(\times\)500 steps; batch 512; LR \(2\!\times\!10^{-3}\!\to\!5\!\times\!10^{-5}\); weight decay 0.01; dropout 0.1; death oversampling \(5\times\). & Max. death F1 (epoch 139). \\
    BC & Three seeds; AdamW; 50 epochs/seed; batch 512; LR \(10^{-3}\); weight decay \(10^{-4}\); positive jump weight 1.5. & Min. loss (10; 10; 11). \\
    PPO & Three seeds; Adam; 15,000 iterations/seed; 512 rollouts/iteration; four update epochs; minibatch 512; LR \(10^{-4}\); entropy/critic weights 0.01/0.5; ratio clip 0.2; max grad. norm 0.5. & Max. survival (12,420; 5,090; 12,280). \\
    \bottomrule
  \end{tabular}
\end{table}

The retained full runs took 3.19 A100 hours for the tokenizer, 4.20 A100 hours for the dynamics model, and 17.41 H200 hours for the seed-43 \(H=45\) controller, including BC and 15,000 PPO iterations.
Thus the retained pipeline required 24.80 hours of sequential training while occupying one GPU at a time---roughly one day end to end.
The matched spatial-only controller took 17.12 H200 hours under the same iteration budget.

\paragraph{Checkpoint and live evaluation.}
PPO checkpoint selection uses survival on 512 fixed latent development contexts every ten iterations.
All live checkpoints are then frozen and act deterministically from pixels alone; Auto-Retry standardizes restarts but provides no policy input.
For each seed, we compare PPO with the exact BC checkpoint that initialized it over 25 scored attempts on \emph{Stereo Madness}, \emph{Back on Track}, and \emph{Polargeist}.
BC and PPO attempts are separate samples paired only by lineage and run until detected death.
Background appearance cycles independently of level progress, while capture and dispatch phases vary near timing boundaries; repetitions measure this pixel-level deployment variability.
The protocol also covers \emph{Stereo Madness Copy}, which begins at Level~1's first ship section, and post-freeze community layout \emph{Stereo INSANE Nerfed}, a held-out test within supported mechanics.
To validate the earlier 30-FPS evaluations, a cadence check compares the identical frozen PPO checkpoint on Stereo Madness Copy over 25 optimized-path attempts at 60 FPS and 20 earlier diagnostic-path attempts at 30 FPS.

The endpoint is frames survived, except the cross-cadence probe uses wall time.
PPO-minus-BC intervals independently resample each frozen policy's 25 attempts 50,000 times and characterize deployment variability; the three seed differences provide replication.

A post-freeze probe retrains seed 43 from the same BC parent with \(H=20\) for 30,000 iterations (15.61 H200 hours), versus \(H=45\) for 15,000 (17.41 hours), and scores 10 versus 25 attempts per layout.
Because horizon and iteration count co-vary, this is a roughly compute-matched shorter-rollout probe, not an isolated ablation.

A second seed-43 probe keeps the \(H=45\) recipe and 15,000-iteration budget but removes \(\mathbf h_t\) from controller training and inference.
It scores the resulting BC and PPO checkpoints over 10 attempts on the three official levels and held-out layout; comparisons with the 25-attempt temporal-state reference independently bootstrap the two frozen-policy samples.

\paragraph{Latency and continuation protocols.}
Live latency covers capture through dispatch over 5,000 optimized-path frames.
The qualitative intervention starts immediately before a spike from one encoded four-frame prefix; greedy branches differ only in their first action and then idle.
Quantitative diagnostics re-encode the unaugmented development trajectories, yielding 22,745 next-frame windows from 413 usable trajectories.
The paired intervention flips only the final context action and bootstraps trajectories 5,000 times; autoregressive diagnostics greedily feed predictions back while replaying recorded actions.
The standard cohort has 1,024 starts from 153 trajectories; the exploratory 200-step cohort has 256 starts from the only four sufficiently long trajectories.

\paragraph{Targeted dynamics-loss ablations.}
We compare the reported structured-SLS+CPC dynamics model with three single-training-run variants: uniform label smoothing with the same \(\varepsilon=0.1\), no label smoothing, and structured SLS without CPC.
They retain the tokenizer, corpus, split, architecture, seed 42, optimization, and maximum-development-death-F1 selection rule; all checkpoints share the exact evaluation windows and rollout starts.

\paragraph{External transition-latency protocol.}
We also time the native imagined transition of DashVMC and released Pong checkpoints of IRIS and DIAMOND on the same RTX 2060.
Each runs in a fresh process at batch one with FP32 eager execution for 30 warm-ups and two repetitions of 100 synchronized transitions; differing native interfaces make this a cost, not quality, comparison.

\FloatBarrier
\section{Results}
\label{sec:results}

\subsection{Policy refinement in the live game}
\label{sec:gd-results}

The central question is whether policy improvement learned only through frozen-model rollouts transfers beyond imitation to the live game.
The selected tokenizer reconstructs development edge maps at 34.10\,dB PSNR, and the dynamics model reaches 29.58\% visual-token accuracy.
Deaths comprise 1.80\% of its 22,745 development windows; at threshold 0.5 the death head obtains 0.721 precision, 0.883 recall, 0.794 F1, 0.994 AUROC, and 0.050 expected calibration error.
Because this stratum selected the dynamics checkpoint, these are in-distribution development diagnostics rather than untouched test results.
These component metrics establish a usable but imperfect imagined environment; the live comparison below tests whether it nevertheless teaches better actions.

\begin{table}[H]
  \centering
  \footnotesize
  \setlength{\tabcolsep}{4pt}
  \caption{Live survival on all five layouts. No-op entries are mean frames $\pm$ standard deviation over 10 attempts and are shared across seeds. BC/PPO entries use 25 attempts; $\Delta$ is PPO minus its exact parent BC [95\% attempt-level bootstrap interval]. The first three are official levels, Stereo Madness Copy begins in ship form, and Stereo INSANE Nerfed is held out.}
  \label{tab:live-controls}
  \begin{tabular}{@{}>{\raggedright\arraybackslash}p{0.18\textwidth}lccc@{}}
    \toprule
    Level & Policy & Seed 43 & Seed 44 & Seed 45 \\
    \midrule
    Stereo Madness & No-op & \multicolumn{3}{c}{shared: $46.2\pm0.7$} \\
       & BC & $120.2\pm91.8$ & $130.6\pm76.0$ & $156.0\pm100.2$ \\
       & PPO & $\boldsymbol{280.1\pm23.3}$ & $\boldsymbol{280.4\pm33.5}$ & $\boldsymbol{308.8\pm68.6}$ \\
       & $\Delta$ [95\% CI] & $159.9\ [122.7,195.4]$ & $149.8\ [116.8,180.6]$ & $152.8\ [106.7,199.8]$ \\
    \addlinespace
    Back on Track & No-op & \multicolumn{3}{c}{shared: $63.4\pm0.7$} \\
       & BC & $123.5\pm81.5$ & $111.1\pm60.3$ & $120.2\pm69.6$ \\
       & PPO & $\boldsymbol{260.0\pm55.5}$ & $\boldsymbol{195.9\pm126.5}$ & $\boldsymbol{209.4\pm130.0}$ \\
       & $\Delta$ [95\% CI] & $136.5\ [98.8,174.4]$ & $84.8\ [31.9,139.6]$ & $89.2\ [34.6,147.2]$ \\
    \addlinespace
    Polargeist & No-op & \multicolumn{3}{c}{shared: $41.5\pm0.7$} \\
       & BC & $52.4\pm16.7$ & $49.0\pm16.5$ & $46.2\pm9.2$ \\
       & PPO & $\boldsymbol{65.2\pm33.4}$ & $\boldsymbol{68.3\pm44.1}$ & $\boldsymbol{63.1\pm32.9}$ \\
       & $\Delta$ [95\% CI] & $12.8\ [-0.2,28.6]$ & $19.4\ [3.0,39.0]$ & $16.8\ [5.7,31.5]$ \\
    \midrule
    Stereo Madness Copy & No-op & \multicolumn{3}{c}{shared: $135.0\pm0.0$} \\
       & BC & $366.8\pm172.8$ & $285.7\pm153.5$ & $299.4\pm175.2$ \\
       & PPO & $\boldsymbol{435.6\pm196.1}$ & $\boldsymbol{495.8\pm137.5}$ & $\boldsymbol{546.8\pm94.1}$ \\
       & $\Delta$ [95\% CI] & $68.8\ [-33.3,166.1]$ & $210.1\ [127.9,285.5]$ & $247.4\ [167.6,320.7]$ \\
    \addlinespace
    Stereo INSANE Nerfed & No-op & \multicolumn{3}{c}{shared: $46.1\pm0.8$} \\
       & BC & $104.8\pm72.1$ & $102.1\pm64.0$ & $97.8\pm57.9$ \\
       & PPO & $\boldsymbol{290.4\pm48.6}$ & $\boldsymbol{342.8\pm80.8}$ & $\boldsymbol{266.7\pm128.9}$ \\
       & $\Delta$ [95\% CI] & $185.5\ [151.1,217.6]$ & $240.6\ [200.8,280.2]$ & $168.8\ [115.4,223.9]$ \\
    \bottomrule
  \end{tabular}
\end{table}

In Table~\ref{tab:live-controls}, BC exceeds no-op everywhere and PPO improves over its parent in all fifteen seed-level comparisons.
Across the three controller seeds, the paired improvements are $154.2\pm5.2$ frames on \emph{Stereo Madness}, $103.5\pm28.7$ on \emph{Back on Track}, and $16.3\pm3.3$ on \emph{Polargeist}, where the variation is the sample standard deviation across the three seed differences.
Every interval on the first two levels excludes zero; the smaller Polargeist gain is positive for all seeds, but seed 43's interval marginally includes zero.
Polargeist's smaller gain may reflect its yellow-orb mechanic, which requires a second, precisely timed jump input while airborne and is absent from the preceding levels.

On Stereo Madness Copy, seed 43's interval includes zero but seeds 44--45 exclude it (mean gain $175.4\pm94.2$); on held-out Stereo INSANE Nerfed, all exclude zero (mean gain $198.3\pm37.6$).
The Copy is not held out, but exposes strong altitude control from Level~1's first ship section.
The 60-FPS cadence probe on Stereo Madness Copy yields similar observed real-time survival: mean recorded episode wall time is 17.72\,s at 60 FPS versus 17.79\,s at 30 FPS, with a 95\% bootstrap interval of $[-1.04,0.89]$\,s for the mean difference (60 FPS minus 30 FPS).
This limited one-checkpoint, one-layout probe supports cadence transfer there, not equivalence across the primary cube levels.

\paragraph{Runtime temporal-state ablation.}
The matched spatial-only controller gives up 1.11 frames on the fixed dream-selection metric, reaching 29.02/45 versus 30.13/45 for the seed-43 temporal-state reference.

\begin{table}[H]
  \centering
  \footnotesize
  \setlength{\tabcolsep}{3pt}
  \caption{Seed-43 controller-input ablation (mean frames $\pm$ sample SD). Dream is best survival on the same 512 fixed contexts; live results use 25 attempts for the temporal-state reference and 10 for spatial-only.}
  \label{tab:temporal-ablation}
  \begin{tabular}{@{}lrrrrr@{}}
    \toprule
    Controller input & Dream & Stereo & Back & Polargeist & INSANE \textit{Nerfed} \\
    \midrule
    \(\mathbf z_t+\mathbf h_t\) & 30.13 & $280.1\pm23.3$ & $260.0\pm55.5$ & $65.2\pm33.4$ & $290.4\pm48.6$ \\
    \(\mathbf z_t\) only & 29.02 & $277.9\pm19.2$ & $294.9\pm64.4$ & $128.0\pm65.1$ & $248.1\pm77.3$ \\
    \bottomrule
  \end{tabular}
\end{table}

Live performance is largely retained without \(\mathbf h_t\).
This suggests that temporal state contributes little in this predominantly reactive game, where the current grid exposes nearby geometry and avatar position and recent history mainly disambiguates motion such as scrolling and vertical velocity.
A memory-dependent environment would be expected to show a larger drop when \(\mathbf h_t\) is removed.
Although \(\mathbf h_t\) improves dream-selection survival by 1.11 frames, the world model's demonstrated role remains to provide PPO's offline training environment.

\FloatBarrier
\subsection{Live control at game speed}
\label{sec:deployment-results}

The complete perception-to-action loop must finish within 16.67\,ms; at 60 Hz it can revise the binary key state every frame, not press 60 times per second.
While the RTX 2060 also renders the game, 5,000 measurements have mean/median 12.018/12.083\,ms, p95 14.068\,ms, p99 14.913\,ms, and maximum 24.059\,ms.
Eleven frames (0.22\%) exceed the period; on a miss the loop skips its pacing sleep and immediately begins the next capture rather than reusing an old action.
This doubles the temporal resolution of the 30-FPS demonstrations.
Qualitatively, the ship policy uses sustained thrust and release intervals: ship motion remains correctable in flight, whereas a cube jump is largely ballistic until landing and therefore demands sparser, more precise timing.
The smooth, low-duty-cycle behaviour is consistent with PPO's per-step jump penalty, although it does not isolate that penalty's effect.

\FloatBarrier
\subsection{Cost of an imagined step}
\label{sec:transition-latency}

DashVMC's native imagined transition averages 11.05\,ms, compared with 49.30\,ms for DIAMOND and 177.83\,ms for IRIS, corresponding to 4.5$\times$ and 16.1$\times$ lower latency, respectively.
Only DashVMC fits the 16.67-ms 60-FPS period on this hardware.

\begin{table}[H]
  \centering
  \footnotesize
  \setlength{\tabcolsep}{4pt}
  \caption{Batch-1 native imagined-transition latency on an RTX 2060 (synchronized run means in milliseconds). Interfaces differ, so this compares operational cost rather than quality.}
  \label{tab:transition-latency}
  \begin{tabular}{lrrr}
    \toprule
    Model & Run 1 & Run 2 & Mean \\
    \midrule
    DashVMC & 11.04 & 11.06 & \textbf{11.05} \\
    DIAMOND & 50.13 & 48.46 & 49.30 \\
    IRIS & 176.94 & 178.73 & 177.83 \\
    \bottomrule
  \end{tabular}
\end{table}

Parallel grid prediction and latent-space feedback avoid autoregressive token generation and pixel decoding at every imagined step, compounding the savings through PPO.

\FloatBarrier
\subsection{Action sensitivity and autoregressive fidelity}
\label{sec:continuations}

Beyond its 45-step training horizon, the model feeds predicted grids and new actions back into context.
Interactive rollouts reproduce scrolling terrain, camera and height changes, collisions, and avatar transformations; long continuations may remain coherent or eventually repeat, empty, or invent impossible geometry.
Video demonstrations of interactive level continuations are available on the \href{https://tariolle.github.io/dash-vmc/\#dreams}{project page}.

\begin{table}[H]
  \centering
  \footnotesize
  \setlength{\tabcolsep}{4pt}
  \caption{Recorded-action rollout diagnostics. Accuracy and decoder-space PSNR compare predicted with recorded grids; JS compares pooled token marginals. Standard: 1,024 starts from 153 trajectories; extended: 256 starts from four long trajectories.}
  \label{tab:rollout-diagnostics}
  \begin{tabular}{lrrrr}
    \toprule
    Cohort & Horizon & Token acc. $\uparrow$ & PSNR $\uparrow$ & Token JS $\downarrow$ \\
    \midrule
    Standard & 1 & 30.14\% & 32.83 & 0.0037 \\
             & 5 & 19.29\% & 28.25 & 0.0045 \\
             & 10 & 14.06\% & 25.23 & 0.0052 \\
             & 20 & 8.26\% & 22.45 & 0.0079 \\
             & 45 & 2.91\% & 19.99 & 0.0208 \\
    \addlinespace
    Extended & 45 & 3.02\% & 19.78 & 0.0351 \\
             & 100 & 1.38\% & 19.49 & 0.0428 \\
             & 200 & 0.74\% & 19.39 & 0.0733 \\
    \bottomrule
  \end{tabular}
\end{table}

Table~\ref{tab:rollout-diagnostics} is best read relative to screen replacement: scrolling replaces the entire visible screen in roughly 45 frames.
The decline from 32.83\,dB at horizon 1 to 19.99\,dB at 45 therefore marks a transition to fully generated geometry, not only compounding error.
Thereafter PSNR is nearly flat within the extended cohort (19.78/19.49/19.39\,dB at horizons 45/100/200): long rollouts can remain coherent, but beyond one screen of travel they are not reliable counterfactual environments for PPO.

Figure~\ref{fig:continuation} instead isolates action sensitivity: from one fixed context, changing one action makes the jump branch clear the spike while idle collides.
Across all 22,745 development transitions, the factual action has lower next-frame negative log-likelihood than its flipped counterpart in 69.52\% of contexts.
The trajectory-mean advantage is 0.165 nats per visual token (95\% episode-bootstrap interval [0.150, 0.181]); factual-action token accuracy is 29.58\% versus 27.78\% after flipping, and 11.04\% of argmax token predictions change.

\begin{figure}[H]
  \centering
  \ifdashvmcworkshop
    \includegraphics[width=\textwidth]{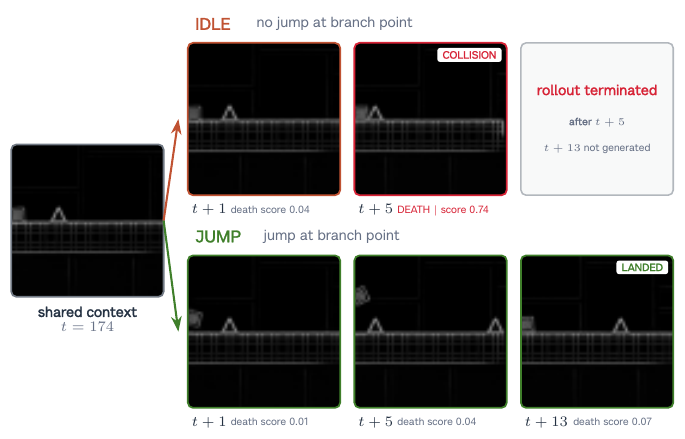}
  \else
    \includegraphics[width=0.70\textwidth]{figures/continuation.pdf}
  \fi
  \caption{Action-conditioning diagnostic from one shared four-frame context. The rows are aligned at \(t+1\) and \(t+5\): changing only the first future action makes the idle branch collide and terminate, while the jump branch clears the spike and lands by \(t+13\).}
  \label{fig:continuation}
\end{figure}

\begin{table}[H]
  \centering
  \footnotesize
  \setlength{\tabcolsep}{4pt}
  \caption{Matched dynamics-loss ablations on fixed development data. \textbf{Ours} uses structured SLS and CPC; other row names denote the changed component. Action advantage is the episode-mean factual-versus-flipped NLL difference. Rollout columns average horizons 1/5/10/20/45. Each row is one run; bold is best.}
  \label{tab:loss-ablation}
  \begin{tabular}{@{}lrrrrr@{}}
    \toprule
    Variant
      & Death F1 $\uparrow$
      & Action adv. $\uparrow$
      & Mean acc. (\%) $\uparrow$
      & Mean PSNR $\uparrow$
      & Mean JS $\downarrow$ \\
    \midrule
    \textbf{Ours}
      & .794 & .165 & \textbf{14.93} & \textbf{25.75} & \textbf{.0084} \\
    No SLS
      & .808 & \textbf{.196} & 14.33 & 25.65 & .0115 \\
    No CPC
      & .787 & .148 & 13.89 & 25.56 & .0104 \\
    Uniform LS
      & \textbf{.826} & .156 & 12.90 & 25.41 & .0141 \\
    \bottomrule
  \end{tabular}
\end{table}

Table~\ref{tab:loss-ablation} averages each rollout curve over the five horizons; Ours leads accuracy and PSNR while minimizing token-marginal JS, and removing CPC worsens every endpoint.
Removing SLS improves death F1 and factual-action advantage but reduces all three rollout aggregates, whereas uniform smoothing attains the highest death F1 but the weakest rollout profile.
These single-run diagnostics support CPC and associate FSQ-structured smoothing with better rollout fidelity, without establishing training variance or downstream policy effects.

\FloatBarrier
\paragraph{Rollout horizon.}

The roughly compute-matched seed-43 \(H=20\) probe changes macro-average live survival by \(+1.9\%\) relative to \(H=45\), with a 95\% stratified-bootstrap interval of \([-8.1,13.5]\%\); every per-layout interval includes zero.
Thus 20-step dreams preserve observed performance and show that late \(H=45\) states are not necessary for this run, but the single seed, unequal attempts, and co-varying iteration count preclude an isolated horizon claim.

\FloatBarrier
\section{Limitations}
\label{sec:limitations}

DashVMC studies one deterministic binary-action game, with one post-freeze layout within known mechanics.
Repeated level prefixes make model diagnostics in-distribution; three controller seeds share one tokenizer and dynamics checkpoint, measuring controller rather than end-to-end uncertainty.
A matched seed-43 spatial-only probe finds no consistent live deficit after removing \(\mathbf h_t\), but its 10-attempt comparison does not establish equivalence; no direct frame-stack controller isolates the value of discrete visual tokens.
Death calibration covers only recorded states, leaving recursive reward exploitation unquantified.
Loss ablations are single-run model diagnostics; controller-input and \(H=20\) probes are single-seed, and the latter is only roughly compute matched.
One-GPU access constrained multi-seed ablations and scaling curves, while wall-clock live evaluation limited deployment tests.
Finally, 60-FPS behavioural equivalence is probed on one checkpoint and layout, and reported speed depends on the RTX 2060 and compact edge representation.

\FloatBarrier
\section{Conclusion}
\label{sec:conclusion}

DashVMC turns a fixed archive of roughly two hours of action-labelled gameplay into an interactive, action-conditioned predictive model of the game.
Its latent rollouts form a learned environment for offline model-based reinforcement learning: PPO improves BC-initialized controllers without additional live interaction, and the resulting policies consistently outlive their BC parents after zero-shot transfer to the non-paused game.
The deployed pipeline meets the game's 60-Hz timing budget, while the spatial-only ablation shows that the learned dynamics can enable policy improvement without remaining in the live control path.
Under data scarcity, the world model thus acts as an interface between past experience and future decisions without prescribing how it must be used: DashVMC uses that interface for imagined policy optimization, while planning-based agents could instead query it at evaluation time.

\ifdashvmcworkshop
  \begin{ack}
    \dashvmcacknowledgments
  \end{ack}
\else
  \raggedbottom
  \clearpage
  \section*{Acknowledgments}
  \dashvmcacknowledgments
\fi

\bibliographystyle{\dashvmcbibliographystyle}
\bibliography{references}

\begin{thebibliography}{17}
\providecommand{\natexlab}[1]{#1}
\providecommand{\url}[1]{\texttt{#1}}
\expandafter\ifx\csname urlstyle\endcsname\relax
  \providecommand{\doi}[1]{doi: #1}\else
  \providecommand{\doi}{doi: \begingroup \urlstyle{rm}\Url}\fi

\bibitem[LeCun(2022)]{lecun2022path}
Yann LeCun.
\newblock A path towards autonomous machine intelligence.
\newblock OpenReview, 2022.
\newblock URL \url{https://openreview.net/pdf?id=BZ5a1r-kVsf}.
\newblock Version 0.9.2.

\bibitem[Ha and Schmidhuber(2018)]{ha2018world}
David Ha and J{\"u}rgen Schmidhuber.
\newblock Recurrent world models facilitate policy evolution.
\newblock In \emph{Advances in Neural Information Processing Systems},
  volume~31, 2018.

\bibitem[Schulman et~al.(2017)Schulman, Wolski, Dhariwal, Radford, and
  Klimov]{schulman2017ppo}
John Schulman, Filip Wolski, Prafulla Dhariwal, Alec Radford, and Oleg Klimov.
\newblock Proximal policy optimization algorithms.
\newblock \emph{arXiv preprint arXiv:1707.06347}, 2017.

\bibitem[Micheli et~al.(2023)Micheli, Alonso, and Fleuret]{micheli2023iris}
Vincent Micheli, Eloi Alonso, and Fran{\c{c}}ois Fleuret.
\newblock Transformers are sample-efficient world models.
\newblock In \emph{International Conference on Learning Representations}, 2023.

\bibitem[Micheli et~al.(2024)Micheli, Alonso, and
  Fleuret]{micheli2024deltairis}
Vincent Micheli, Eloi Alonso, and Fran{\c{c}}ois Fleuret.
\newblock Efficient world models with context-aware tokenization.
\newblock In \emph{Proceedings of the 41st International Conference on Machine
  Learning}, volume 235 of \emph{Proceedings of Machine Learning Research},
  pages 35623--35638. PMLR, 2024.

\bibitem[Alonso et~al.(2024)Alonso, Jelley, Micheli, Kanervisto, Storkey,
  Pearce, and Fleuret]{alonso2024diamond}
Eloi Alonso, Adam Jelley, Vincent Micheli, Anssi Kanervisto, Amos Storkey, Tim
  Pearce, and Fran{\c{c}}ois Fleuret.
\newblock Diffusion for world modeling: Visual details matter in atari.
\newblock In \emph{Advances in Neural Information Processing Systems},
  volume~37, 2024.

\bibitem[Hafner et~al.(2025)Hafner, Pasukonis, Ba, and
  Lillicrap]{hafner2025dreamerv3}
Danijar Hafner, Jurgis Pasukonis, Jimmy Ba, and Timothy Lillicrap.
\newblock Mastering diverse control tasks through world models.
\newblock \emph{Nature}, 640:\penalty0 647--653, 2025.
\newblock \doi{10.1038/s41586-025-08744-2}.

\bibitem[Burchi and Timofte(2025)]{burchi2025twister}
Maxime Burchi and Radu Timofte.
\newblock Learning transformer-based world models with contrastive predictive
  coding.
\newblock In \emph{International Conference on Learning Representations}, 2025.

\bibitem[Li and Rafferty(2017)]{li2017geometrydash}
Ted Li and Sean Rafferty.
\newblock Playing {Geometry Dash} with convolutional neural networks.
\newblock Stanford CS231N course report, 2017.
\newblock URL \url{https://cs231n.stanford.edu/reports/2017/pdfs/605.pdf}.

\bibitem[Mentzer et~al.(2024)Mentzer, Minnen, Agustsson, and
  Tschannen]{mentzer2024fsq}
Fabian Mentzer, David Minnen, Eirikur Agustsson, and Michael Tschannen.
\newblock Finite scalar quantization: {VQ-VAE} made simple.
\newblock In \emph{International Conference on Learning Representations}, 2024.

\bibitem[Lin et~al.(2026)Lin, Li, Niu, Gong, Ge, Lin, Zheng, Zhang, Yang,
  Zhong, Bo, and Yuan]{lin2026ifsq}
Bin Lin, Zongjian Li, Yuwei Niu, Kaixiong Gong, Yunyang Ge, Yunlong Lin,
  Mingzhe Zheng, JianWei Zhang, Miles Yang, Zhao Zhong, Liefeng Bo, and
  Li~Yuan.
\newblock i{FSQ}: Improving {FSQ} for image generation with 1 line of code.
\newblock \emph{arXiv preprint arXiv:2601.17124}, 2026.

\bibitem[Hester et~al.(2018)Hester, Vecerik, Pietquin, Lanctot, Schaul, Piot,
  Horgan, Quan, Sendonaris, Osband, Dulac-Arnold, Agapiou, Leibo, and
  Gruslys]{hester2018deep}
Todd Hester, Matej Vecerik, Olivier Pietquin, Marc Lanctot, Tom Schaul, Bilal
  Piot, Dan Horgan, John Quan, Andrew Sendonaris, Ian Osband, Gabriel
  Dulac-Arnold, John Agapiou, Joel~Z. Leibo, and Audrunas Gruslys.
\newblock Deep q-learning from demonstrations.
\newblock In \emph{AAAI Conference on Artificial Intelligence}, volume~32,
  2018.

\bibitem[Baker et~al.(2022)Baker, Akkaya, Zhokhov, Huizinga, Tang, Ecoffet,
  Houghton, Sampedro, and Clune]{baker2022vpt}
Bowen Baker, Ilge Akkaya, Peter Zhokhov, Joost Huizinga, Jie Tang, Adrien
  Ecoffet, Brandon Houghton, Raul Sampedro, and Jeff Clune.
\newblock Video pretraining ({VPT}): Learning to act by watching unlabeled
  online videos.
\newblock \emph{arXiv preprint arXiv:2206.11795}, 2022.

\bibitem[Xia et~al.(2025)Xia, Lu, Li, Xu, and Chen]{xia2025grwm}
Zaishuo Xia, Yukuan Lu, Xinyi Li, Yifan Xu, and Yubei Chen.
\newblock Cloning deterministic worlds: The critical role of latent geometry in
  long-horizon world models.
\newblock \emph{arXiv preprint arXiv:2510.26782}, 2025.

\bibitem[van~den Oord et~al.(2018)van~den Oord, Li, and Vinyals]{oord2018cpc}
Aaron van~den Oord, Yazhe Li, and Oriol Vinyals.
\newblock Representation learning with contrastive predictive coding.
\newblock \emph{arXiv preprint arXiv:1807.03748}, 2018.

\bibitem[Lin et~al.(2017)Lin, Goyal, Girshick, He, and
  Doll{\'a}r]{lin2017focal}
Tsung-Yi Lin, Priya Goyal, Ross Girshick, Kaiming He, and Piotr Doll{\'a}r.
\newblock Focal loss for dense object detection.
\newblock In \emph{International Conference on Computer Vision}, pages
  2980--2988, 2017.

\bibitem[Schulman et~al.(2015)Schulman, Moritz, Levine, Jordan, and
  Abbeel]{schulman2015gae}
John Schulman, Philipp Moritz, Sergey Levine, Michael Jordan, and Pieter
  Abbeel.
\newblock High-dimensional continuous control using generalized advantage
  estimation.
\newblock \emph{arXiv preprint arXiv:1506.02438}, 2015.

\end{thebibliography}

\end{document}